\def\year{2023}\relax
\documentclass[letterpaper]{article} 
\usepackage{aaai22}  
\usepackage{times}  
\usepackage{helvet}  
\usepackage{courier}  
\usepackage[hyphens]{url}  
\usepackage{graphicx} 
\usepackage{amssymb}
\usepackage{booktabs}
\usepackage{nameref}
\usepackage[utf8]{inputenc}
\usepackage{natbib}  
\usepackage{caption} 
\DeclareCaptionStyle{ruled}{labelfont=normalfont,labelsep=colon,strut=off} 
\usepackage{caption}
\usepackage{algorithm}
\usepackage{algorithmic}
\usepackage{bm}
\usepackage{color}
\usepackage{soul}

\usepackage{tabularx}
\usepackage{multirow}
\usepackage{amsmath}
\usepackage{newfloat}
\usepackage{listings}
\floatstyle{ruled}
\newfloat{listing}{tb}{lst}{}
\floatname{listing}{Listing}
\title{MDM: Molecular Diffusion Model for 3D Molecule Generation}
\author{
Lei Huang,\footnote{This work is done when L. Huang works as an intern in Tencent AI Lab.}\textsuperscript{\rm1 \rm2}
Hengtong Zhang,\footnote{Correspondence to: H. Zhang. } \textsuperscript{\rm1}
Tingyang Xu,\textsuperscript{\rm1}
Ka-Chun Wong \textsuperscript{\rm2}
}
\affiliations {
\textsuperscript{\rm 1}Tencent AI Lab,
\textsuperscript{\rm 2}City University of Hong Kong \\
lhuang93-c@my.cityu.edu.hk,
htzhang.work@gmail.com, \\
tingyangxu@tencent.com,
kc.w@cityu.edu.hk
}

\usepackage{bibentry}

\begin{document}

\maketitle

\begin{abstract}
Molecule generation, especially generating 3D molecular geometries from scratch (i.e., 3D \textit{de novo} generation), has become a fundamental task in drug designs. Existing diffusion-based 3D molecule generation methods could suffer from unsatisfactory performances, especially when generating large molecules. At the same time, the generated molecules lack enough diversity. This paper proposes a novel diffusion model to address those two challenges. 

First, interatomic relations are not in molecules' 3D point cloud representations. Thus, it is difficult for existing generative models to capture the potential interatomic forces and abundant local constraints. 
To tackle this challenge, we propose to augment the potential interatomic forces and further involve dual equivariant encoders to encode interatomic forces of different strengths.
Second, existing diffusion-based models essentially shift elements in geometry along the gradient of data density. Such a process lacks enough exploration in the intermediate steps of the Langevin dynamics. To address this issue, we introduce a distributional controlling variable in each diffusion/reverse step to enforce thorough explorations and further improve generation diversity.

Extensive experiments on multiple benchmarks demonstrate that the proposed model significantly outperforms existing methods for both unconditional and conditional generation tasks. We also conduct case studies to help understand the physicochemical properties of the generated molecules.
\end{abstract}

\section{Introduction}

\textit{De novo} molecule generation, which automatically generates valid chemical structures with desirable properties, has become a crucial task in the domain of drug discovery. 
However, given the huge diversity of atom types and chemical bonds, the manually daunting task in proposing valid, unique, and property-restricted molecules is extraordinarily costly.  
To tackle such a challenge, a multitude of generative machine learning models~\cite{zang2020moflow,satorras2021en,gebauer2019symmetry,hoogeboom2022equivariant}, which automatically generate molecular geometries (i.e., 2D graphs or 3D point clouds) from scratch, has been proposed in the past decade. 

Among those studies, 3D molecule generation has become an emerging research topic due to its capability of directly generating 3D coordinates of atoms, which are important in determining molecules' physical-chemical properties. Early studies on 3D molecule generation adopt auto-regressive models such as normalized flow~\cite{satorras2021en} to determine the types and 3D positions of atoms one by one. Nevertheless, these models suffer from deviation accumulations, especially when invalid structures are generated in the early steps (i.e. initial condition vulnerability). Such a major drawback leads to unsatisfactory generation results in terms of molecule validity and stability. Later, inspired by the success of diffusion models, \cite{hoogeboom2022equivariant} proposed a cutting-edge diffusion based 3D generation model that significantly improves the validity of generated molecules. The diffusion based generation model~\cite{hoogeboom2022equivariant} defines a Markov chain of diffusion steps to add random noises to 3D molecule geometries and then learns a reverse process to construct desired 3D geometries step-by-step.

Nonetheless, diffusion-based 3D generation models still suffer from two non-negligible drawbacks: First, unlike 2D generation, in which chemical bonds are represented as graph edges, molecular geometries in 3D generation are represented as point clouds~\cite{satorras2021en,gebauer2019symmetry,hoogeboom2022equivariant}. Hence, it is difficult for 3D generative models to capture the abundant local constraint relations between adjacent atoms with no explicit indications for chemical bonds. Such a significant drawback leads to unsatisfactory performance on \emph{datasets with large molecules}, for instance, the GEOM dataset~\cite{axelrod2022geom} with average 46 atoms per molecule. 

Moreover, training diffusion models is essentially equivalent to a denoising score matching process with Langevin dynamics as existing literature~\cite{ho2020denoising,song2021scorebased} suggests, in which the elements in a geometry (i.e., points in a 3D point cloud) shift along the gradient of data density at each timestep. Thus, the generation dynamics by given fixed initialized noise may concentrate around a common trajectory and leads to similar generation results, even with the standard Gaussian noise compensations in the sampling process. Such a phenomenon hurts the diversity of generated molecules in practice.

In this paper, we propose a novel model named MDM (\textbf{M}olecular \textbf{D}iffusion \textbf{M}odel) to tackle these drawbacks. First, we propose to treat atoms pairs with atomic spacing below some threshold\footnote{The distance value threshold varies for different types of bonds.} as covalently bonded since chemical bonds can dominate the interatomic force when two atoms are close enough to each other. We can thus construct augmented bond-like linkages between adjacent atoms. On the other hand, for the atoms pairs with atomic spacing above some thresholds, the van der Waals force dominate the interatomic force. Those two types suggest different strengths between atoms and thus should be treated distinctly. Given such intuition, we deploy separated equivariant networks to explicitly model the destination between these two kinds of inter-atom bonds. 

Moreover, to enhance the diversity of molecular generation, we introduce latent variables, interpreted as controlling representations in each diffusion/reverse step of a diffusion model. Thus, each diffusion/reverse step is conditioned to a distributional (e.g. Gaussian) representation that can be effectively explored. In the generation\footnote{Or sampling phase in the context of the diffusion model.} phase, by sampling from the underlying distribution of the variable in each step, we can enforce thorough explorations to generate diverse 3D molecule geometries.

Experiments on two  molecule datasets (i.e., QM9~\cite{ramakrishnan2014quantum} and GEOM~ \cite{axelrod2022geom}) demonstrate that the proposed MDM outperforms the state-of-the-art model EDM~\cite{hoogeboom2022equivariant} by a wide margin, especially on the drug-like GEOM dataset that consists of molecules with a large number of atoms (46 atoms on average compared with 18 in QM9).
Remarkably, the uniqueness and novelty metric, \textcolor{black}{which characterizes the diversity of generative molecules}, is improved by $4.1\%$ to $31.4\%$ compared with EDM.
We also present studies on targeted molecular generation tasks to show that the proposed model is capable of generating molecules with chosen properties without scarifying the validity and stability of generated molecules.

\section{Related work}
Deep generative models have recently exhibited their effectiveness in modeling the density of real-world molecule data for molecule design and generation. Various methods firstly consider the molecule generation in a 2D fashion. Some methods \cite{dai2018syntaxdirected, gomez2018automatic, grisoni2020bidirectional} utilize sequential models such as RNN to generate SMILES \cite{weininger1988smiles} strings of molecules while other models focus on molecular graphs whose atoms and chemical bonds are represented by nodes and edges. Generally, these methods incorporate the variational autoencoder (VAE)-based models \cite{jin2018junction}, generative adversarial network (GAN) \cite{de2018molgan} and normalizing flows \cite{zang2020moflow, luo2021graphdf} to generate the atom types and the corresponding chemical bonds in one-shot or auto-regressive manners. Although these studies are able to generate valid and novel molecule graphs, they ignore the 3D structure information of molecules which is crucial for determining molecular properties.

Recently, generating molecules in 3D space has gained a lot of attention. 
For instance, G-Schnet \cite{gebauer2019symmetry} employs an auto-regressive process equipped with Schnet \cite{schutt2017schnet} in which atoms and bonds are sampled iteratively. E-NF \cite{garcia2021n} instead utilizes a one-shot framework based on an equivariant normalizing flow to generate atom types and coordinates at one time. Recently, inspired by the success of diffusion models \cite{sohl2015deep} in various tasks \cite{ho2020denoising, song2021denoising, kong2021diffwave}, \cite{hoogeboom2022equivariant} adopts the diffusion model to generate novel molecules in 3D space. However, it only utilizes the fully connected adjacent matrix thus ignoring the intrinsic topology of the molecular graph. 

Apart from molecule generation, the task discussed in this paper is also related to conformation prediction~\cite{mansimov2019molecular, kohler2020equivariant, xu2021end, guan2022energyinspired}. Although both molecule generation and conformation prediction output 3D molecule geometries, the settings of these two tasks are different. The former can directly generate a complete molecule while the latter additionally requires molecular graphs as inputs and only outputs atom coordinates.

\section{Preliminaries}

\textbf{Notation.} Let $\mathcal{G} = (A,R)$ denote the 3D molecular geometry as  where $A \in \{0,1\}^{n \times f}$ denotes the atom features, including atom types and atom charges. $R \in \mathbb{R}^{n \times 3}$ denotes the atom coordinates. 


\begin{figure*}[ht]
    \centering
    \includegraphics[height=0.4\textwidth]{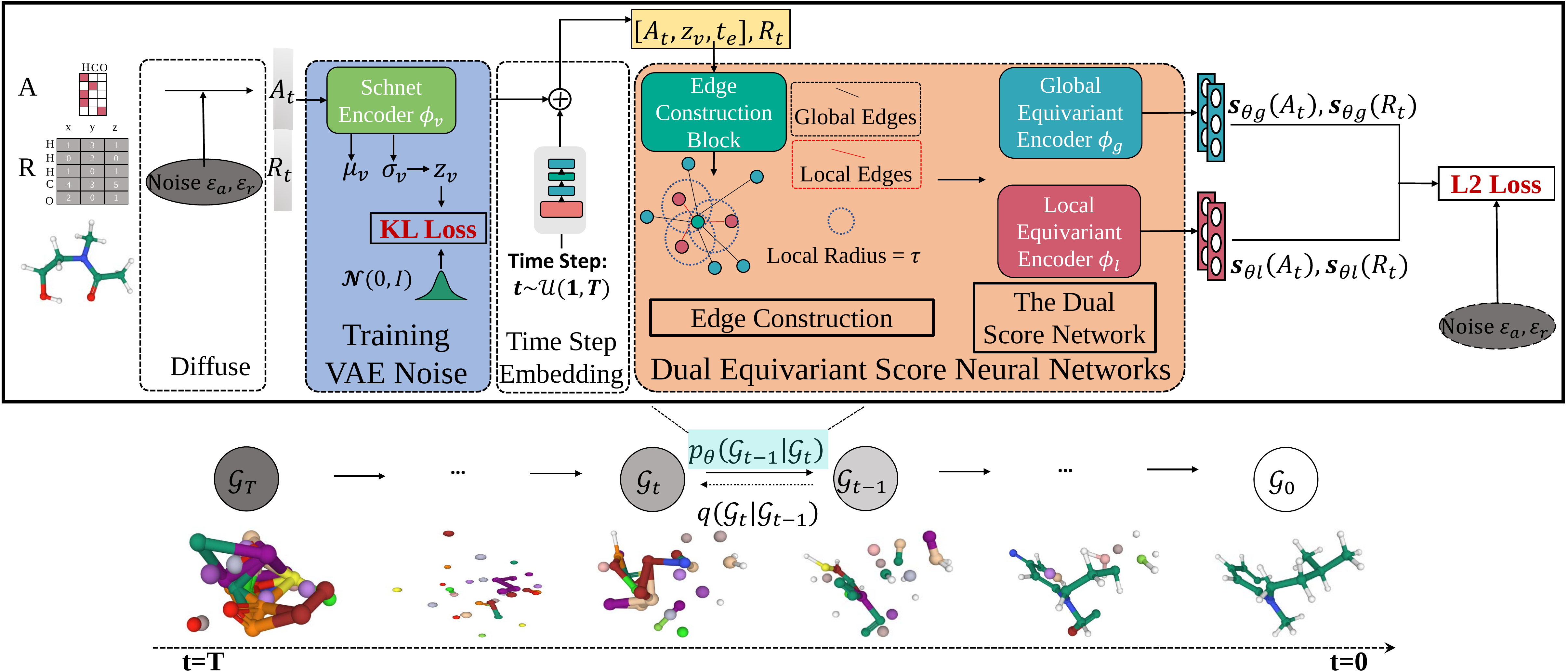}
    \caption{Overview of the training process of the proposed model MDM. The model would train each molecule which includes the atom features and atom coordinates with a stochastic time step. For reverse process, the final molecule is generated by denoising the initial state $\mathcal{G}_T \sim \mathcal{N}(0, I) $ gradually with the Markov kernels $p_\theta(\mathcal{G}_{t-1} \mid \mathcal{G}_t)$. Symmetrically, the diffusion process is achieved by adding the noise with the posterior distribution $q(\mathcal{G}_{t} \mid \mathcal{G}_{t-1})$ until the molecule is degenerated into the white noise when the time step is large enough. It is also should be noted that the global and local equivariant encoder have the same structure.}  
    \label{fig:model_flow}
\end{figure*}

\subsection{Diffusion Model}
The diffusion model is formulated as two Markov chains: {\em diffusion process} and {\em reverse process} (a.k.a denoising process). 
In the upcoming paragraphs, we will elaborate these two processes. 
\subsubsection*{Diffusion Process.} \label{sec:Diffusion Process} 
Given the real molecular geometry $\mathcal{G}_0$, the forward diffusion process gradually diffuses the data into a predefined noise distribution with the time setting $1 \dots T$, like the physical phenomenon. The diffusion model is formulated as a fixed Markov chain that gradually adds Gaussian noise to the data with a variance schedule $\beta_1 \dots \beta_T (\beta_t \in (0,1))$:      
\begin{equation}
\label{eq: diffusion process}
\begin{split}
&q\left(\mathcal{G}_{1: T} \mid \mathcal{G}_{0}\right)=\prod_{t=1}^{T}q\left(\mathcal{G}_{t} \mid \mathcal{G}_{t-1}\right), \\
&q\left(\mathcal{G}_{t} \mid \mathcal{G}_{t-1}\right)=\mathcal{N}\left(\mathcal{G}_{t} ; \sqrt{1-\beta_{t}} \mathcal{G}_{t-1}, \beta_{t} I\right),
\end{split}
\end{equation}
where $\mathcal{G}_{t-1}$ is mixed with the Gaussian noise to obtain $\mathcal{G}_{t}$ and $\beta_t$ controls the extent of the mixture. By setting $\bar{\alpha}_{t}=\prod_{s=1}^{t} 1-\beta_{s}$, a delightful property of the diffusion process is achieved that any arbitrary time step, $t$, sampling of the data has a closed-form formulation via a reparameterization trick as:
\begin{equation}
q\left(\mathcal{G}_{t} \mid \mathcal{G}_{0}\right)=\mathcal{N}\left(\mathcal{G}_{t} ; \sqrt{\bar{\alpha}_{t}} \mathcal{G}_{0},\left(1-\bar{\alpha}_{t}\right) I\right).
\end{equation}

With step $t$ gradually rises, the final distribution will be closer to the standard Gaussian distribution because $\sqrt{\bar{\alpha}_{t}} \to 0$ and $(1-\bar{\alpha}_{t}) \to 1$ if $t \to \infty$.

\subsubsection*{Reverse Process. } \label{sec:Reverse Process}
The reverse process aims to learn a process to reverse the diffusion process back to the distribution of the real data. Assume that there exists a reverse process, $q\left(\mathcal{G}_{t-1} \mid \mathcal{G}_{t}\right)$. Then such process could generate valid molecules from a standard Gaussian noise following a Markov chain from $T$ back to $0$ as shown in Figure \ref{fig:model_flow}. 
However, it is hard to estimate such distribution, $q\left(\mathcal{G}_{t-1} \mid \mathcal{G}_{t}\right)$. Hence, a learned Gaussian transitions $p_{\theta}\left(\mathcal{G}_{t-1} \mid  \mathcal{G}_{t}\right)$ is devised to approximate the $q\left(\mathcal{G}_{t-1} \mid  \mathcal{G}_{t}\right)$ at every time step: 
\begin{equation}
\begin{split}
&p_{\theta}\left(\mathcal{G}_{0: T-1} \mid \mathcal{G}_{T}\right)=\prod_{t-1}^{T} p_{\theta}\left(\mathcal{G}_{t-1} \mid \mathcal{G}_{t}\right),\\ &p_{\theta}\left(\mathcal{G}_{t-1} \mid  \mathcal{G}_{t}\right)=\mathcal{N}\left(\mathcal{G}_{t-1} ; \boldsymbol{\mu}_{\theta}\left(\mathcal{G}_{t}, t\right), \sigma_{t}^{2} I\right),
\end{split}
\end{equation}
where $\boldsymbol{\mu}_{\theta}$ denotes the parameterized neural networks to approximate the mean, and $\sigma_{t}^2$ denotes user defined variance.

To learn the $\boldsymbol{\mu}_{\theta}\left(\mathcal{G}_{t}, t\right)$, we adopt the following parameterization of $\boldsymbol{\mu}_{\theta}$ following \citet{ho2020denoising}:
\begin{equation}
\label{eq: mu}
    \boldsymbol{\mu}_{\theta}\left(\mathcal{G}_{t}, t\right)=\frac{1}{\sqrt{1-\beta_{t}}}\left(\mathcal{G}_{t}-\frac{\beta_{t}}{\sqrt{1-\bar{\alpha}_{t}}} \boldsymbol{\epsilon}_{\theta}\left(\mathcal{G}_{t}, t\right)\right),
\end{equation}
where $\boldsymbol{\epsilon}_{\theta}$ is a neural network w.r.t trainable parameters $\theta$. 

From another perspective, the reverse process, that eliminates the noise part of the data added in the diffusion process at each time step, is equivalent to a moving process on the data distribution that initially starts from a low density region to the high density region of the distribution led by the logarithmic gradient. Therefore, the negative eliminated noise part $-\boldsymbol{\epsilon}_{\theta}$ is also regarded as the \emph{(stein) score} \cite{liu2016kernelized}, the logarithmic density of the data point at every time step. This equivalence is also reflected in the previous work \cite{song2021scorebased}. For simplicity, we utilize $\boldsymbol{s}_{\theta}$ for all the related formulas in following sections.

Now we can parameterize $\boldsymbol{\mu}_{\theta}\left(\mathcal{G}_{t}, t\right)$ as:
\begin{equation}
\boldsymbol{\mu}_{\theta}\left(\mathcal{G}_{t}, t\right)=
\frac{1}{\sqrt{1-\beta_{t}}}\left(\mathcal{G}_{t}+\frac{\beta_{t}}{\sqrt{1-\bar{\alpha}_{t}}} \boldsymbol{s}_{\theta}\left(\mathcal{G}_{t}, t\right)\right).  
\end{equation}

The complete sampling process resembles Langevin dynamics with $\boldsymbol{s}_\theta$ as a learned gradient of the data density .

\section{MDM: Molecular Diffusion Model}

In this section, we present our proposed model MDM, a molecular diffusion model. 
As shown in Figure \ref{fig:model_flow}, we design {\em dual equivariant score neural networks} to handle two levels of edges: local edges within a predefined radius to model the intramolecular force such as covalent bonds and global edges to capture van der Waals forces. Furthermore, We introduce a {\em VAE module} inside the diffusion model to produce conditional noise which will avoid determining output of the whole model and improve the generation diversity. Then, we describe how the {\em training phase} and {\em sampling phase} of MDM works. 

\begin{figure*}[tbp]
\begin{minipage}{0.45\textwidth}
    \centering
    \includegraphics[width=\linewidth]{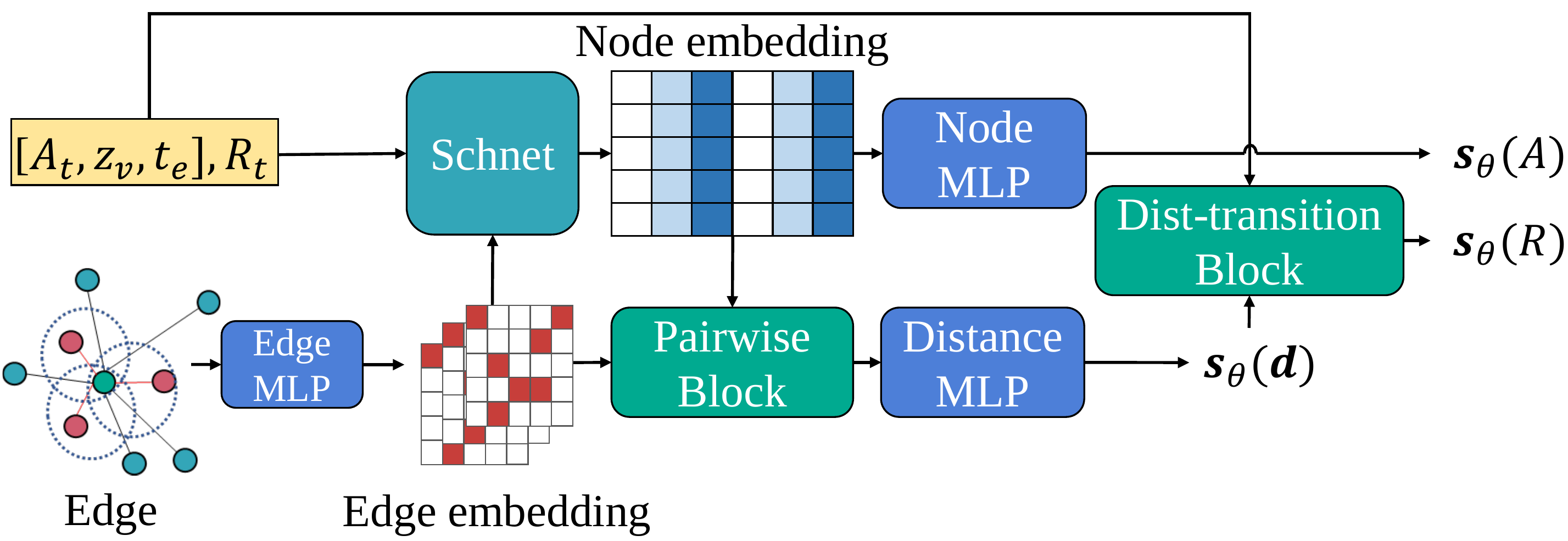}
    \caption{The illustration of the equivariant encoder where Schnet follows the implementation of \citet{schutt2017schnet}.}  
    \label{fig:Equivariant_block}
\end{minipage}
\hfill
\begin{minipage}{0.47\textwidth}
    \centering
    \includegraphics[width=\linewidth]{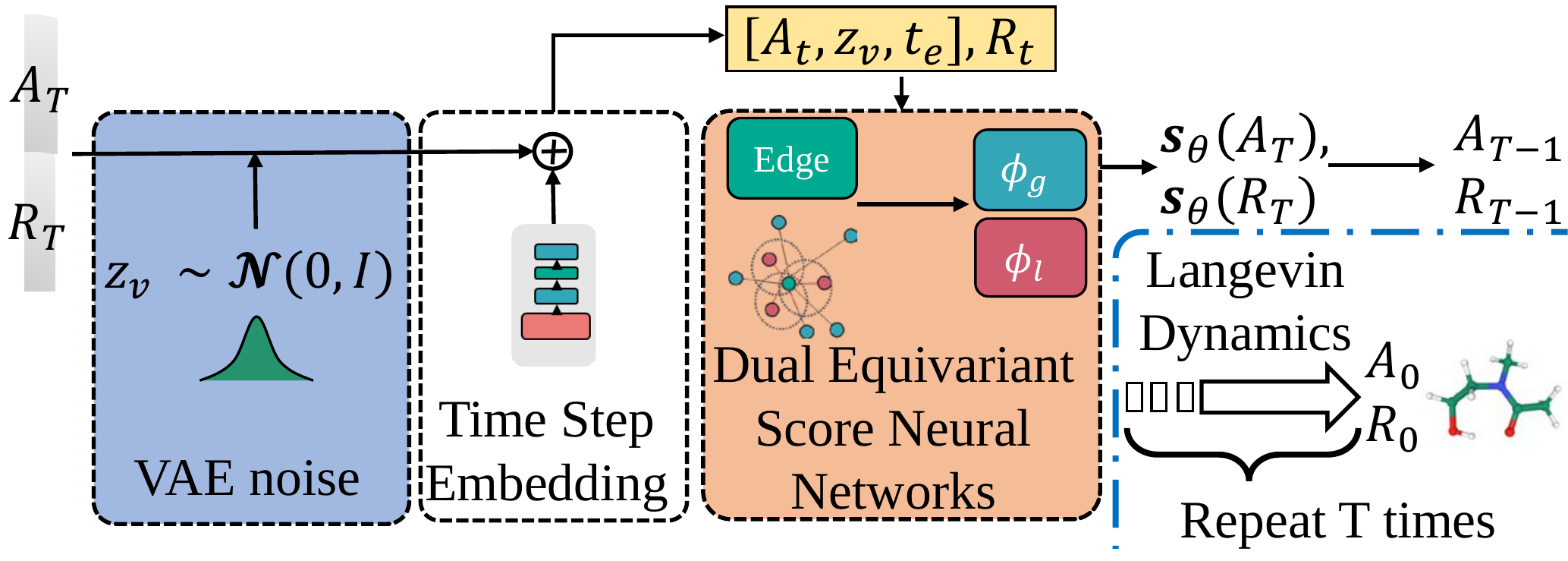}
    \caption{Overview of the sampling process of the proposed model MDM. $A_T$ and $R_T$ are sampled from $\mathcal{N}(0, I)$. }
    \label{fig:model_flow_sampling}
\end{minipage}
\end{figure*}

\subsection*{Dual Equivariant Score Neural Networks}
\label{sec:Equivariant Encoder}

As molecular geometries are roto-translation \emph{invariant}, we should take this property into account when devising the Markov kernels. In essence, \cite{kohler2020equivariant} proposed an equivariant invertible function to transform an \emph{invariant} distribution into another \emph{invariant} distribution. This theorem is also applied to the diffusion model \cite{xu2022geodiff}. If $p(\mathcal{G_T})$ is \emph{invariant} and the neural network $q_\theta$ which learns to parameterize $p(\mathcal{G}_{t-1} \mid \mathcal{G}_{t})$ is \emph{equivariant}, then the margin distribution $p(\mathcal{G})$ is also \emph{invariant}. Therefore, we utilize an \emph{equivariant} Markov kernel to achieve this desired property.

\subsubsection{Edge Construction.} First, we describe how to construct edges at each timestep for the equivariant Markov kernel since the equivariant graph model requires a graph structure as input. Recalls that previous works \cite{kohler2020equivariant, hoogeboom2022equivariant} consider the fully connected edges to feed into the equivariant graph neural network. However, the fully connected edges connect all the atoms and treat the interatomic effects equally but regret the effects of covalent bonds. Therefore, we further define the edges within the radius $\tau$ as local edges to simulate the covalent bonds and the rest of edges in the fully connected edges as global edges to capture the long distance information such as van der Waals force. 

Practically, we set the local radius $\tau$ as $2$\AA because almost all of the chemical bonds are no longer than $2$\AA. The atom features and coordinates with the local edges and global edges are fed into the dual equivariant encoder, respectively. Specifically, the local equivariant encoder models the intramolecular force such as the real chemical bonds via local edges while the global equivariant encoder captures the interactive information among distant atoms such as van der Waals force via global edges. 

\subsubsection{The Equivariant Markov Kernels.}
Both local and global equivariant encoders share the same architecture as Equivariant Markov Kernels. Intuitively, atom features, $A$, are invariant to any rigor transformations on atom coordinates $R$, while $R$ should be equivariant to such transformations. Therefore, we design the equivariant Markov Kernels as Figure \ref{fig:Equivariant_block} to tackle invariance and equivariance of $A$ and $R$, respectively.

First, we consider the invariance of the model for atomic features, $A$. Firstly, we utilize an \textbf{Edge MLP} to obtain edge embeddings as 
\begin{equation}
\mathbf{h}_{e_{ij}} = \text{MLP}(\mathbf{d}_{ij}, e_{ij}),
\end{equation}
where $\mathbf{d}_{ij} = \left\|\mathbf{r}_{i}-\mathbf{r}_{j}\right\|_{2}$ denotes the Euclidean distance between the positions of atom $i$ and atom $j$, and $e_{ij}$ denotes the edge features between $i^{th}$ atom and $j^{th}$ atom. Then we adopt \textbf{Schnet} with $L$ layers to achieve invariance:  
\begin{equation}
\begin{split}
\mathbf{h}^{0}_{iA} = & \text{MLP}(A_{i}), \mathbf{h}^{0}_{iR} = \text{MLP}(R_{i}), \mathbf{h}^{0}_{i} = [\mathbf{h}^{0}_{iA},\mathbf{h}^{0}_{iR}], \\
\mathbf{h}_{i}^{(l+1)}=&\sigma\left(\boldsymbol{W}_{0}^{l} \mathbf{h}_{i}^{(l)}+\sum_{j \in N\left(i\right)} \boldsymbol{W}_{1}^{l} \phi_w\left(\mathbf{d}_{i j}\right) \odot \boldsymbol{W}_{2}^{l} \mathbf{h}_{j}^{(l)}\right),
\end{split}
\end{equation}
where $l\in (0,1,\dots,L)$ indicates the $l^{th}$ layer of Schnet, $\boldsymbol{W}^l$s represent the learning weights. Then, we denote the final outputs of Schnet, $\mathbf{h}_i$, as node embeddings, $\sigma(\cdot)$ denotes non-linear activation such as ReLU, and $\phi_w(\cdot)$ denotes a weight network. Then, the final outputs of Schnet $\mathbf{h}_i=\mathbf{h}^{(L)}_i$ are denoted as node embeddings.

In order to estimate the gradient of log density of atom features, we utilize one-layer \textbf{Node MLP} to map the latent hidden vectors outputted by Schnet to score vectors. 
\begin{equation}
    \boldsymbol{s}_{\theta}(A_{i}) = \text{MLP}(\mathbf{h}_{i}).
\end{equation}

On the other hand, to achieve equivariance for atomic coordinates, $R$, in 3D space, we attempt to decompose them to pairwise distances. Therefore, we concatenate the learned edge information and the product of the end nodes vectors of the same edges as \textbf{Pairwise Block} followed by a \textbf{Distance MLP} to get the gradient of the pairwise distances.
\begin{equation}
    \boldsymbol{s}_{\theta}(\mathbf{d}_{ij}) = \text{MLP}(\left[\mathbf{h}_{i} \cdot \mathbf{h}_{j}, \mathbf{h}_{e_{ij}} \right]).
\end{equation}
Here, we omit $t$ on $\boldsymbol{s}_\theta$ as we only discuss the coordinate part of this function in one time step for simplicity. 

Then, MDM operates a transition, called \textbf{Dist-transition Block}, to integrate the information of score vectors from pairwise distance and atomic coordinates $R$ as follows
\begin{equation}
\mathbf{s}_{\theta}(R_{i})=\sum_{j \in N(i)} \frac{1}{\mathbf{d}_{i j}} \cdot \mathbf{s}_{\theta}(\mathbf{d}_{i j}) \cdot\left(\mathbf{r}_{i}-\mathbf{r}_{j}\right),
\end{equation}
where $\boldsymbol{s}_\theta(d)$ is invariant to translation since it only depends on the symmetry-invariant element $d$ and $\mathbf{r}_{i}-\mathbf{r}_{j}$ is roto-translation equivariance. Thus $\boldsymbol{s}_\theta(R)$ shares the equivariant property.

\subsection*{Enhanced Diversity via Variational Noising} \label{sec:VAE Module}
The diffusion model can be extended to a conditional generation by enforcing the generated samples with the additional given information. Therefore,
we employ variational noising to import an additional noise $z_v$ for conditional generation $p_{\theta}\left(\mathcal{G}_{0: T-1} \mid \mathcal{G}_{T}, z_v \right)$ and improve the diversity. Specifically, we adopt Schnet as the encoder and the subsequent equivariant modules as the decoder. The encoder outputs the mean $\mu_v$ and the standard deviation $\sigma_v$ from which we can obtain the additional noise $z_v$ by the reparameterization trick, $z_v = \mu_v + \sigma_v^2z, z\sim \mathcal{N}(0, I)$. Hence, Eq. \eqref{eq: reverse process} in reverse process of diffusion model becomes
\begin{equation}
\begin{split}\label{eq: reverse process}
&p_{\theta}\left(\mathcal{G}_{0: T-1} \mid \mathcal{G}_{T}, z_v\right)=\prod_{t-1}^{T} p_{\theta}\left(\mathcal{G}_{t-1} \mid \mathcal{G}_{t}, z_v\right),\\ 
&p_{\theta}\left(\mathcal{G}_{t-1} \mid  \mathcal{G}_{t}, z_v\right)=\mathcal{N}\left(\mathcal{G}_{t-1} ; \boldsymbol{\mu}_{\theta}\left(\mathcal{G}_{t}, z_v, t\right), \sigma_{t}^{2} I\right).
\end{split}
\end{equation}
When forwarding the diffusion process, we sample the variational noise $z_v$ from $\mathcal{N}(0, I)$. We also surprisely observe that the performance is improved if we apply the polarized sampling strategy. Empirically, the performance of MDM improves significantly when we sample $z_v$ from a uniform distribution $\mathcal{U}(-1,+1)$.

\begin{algorithm}[t]\small
	\caption{Training Process}
	\label{alg_t}
	\textbf{Input}: The molecular geometry $\mathcal{G}(A,R)$, VAE encoder $\phi_{v}$ global equivariant neural networks $\phi_{g}$, local neural networks $\phi_{l}$
	\begin{algorithmic}[1]
		\REPEAT
		\STATE $\mathbf{a}_{0} \sim q\left(\mathbf{a}_{0}\right);\mathbf{r}_{0} \sim q\left(\mathbf{r_0}\right)$ 
		\STATE $t \sim \mathcal{U}(\{1, \ldots, T\})$,  $\boldsymbol{\epsilon}^a \sim \mathcal{N}(0, I)$,
		$\boldsymbol{\epsilon}^r \sim \mathcal{N}(0, I)$
		\STATE Shift $\boldsymbol{\epsilon}^r$ to zero COM, $\boldsymbol{\epsilon} = \left[\boldsymbol{\epsilon}^a, \boldsymbol{\epsilon}^r \right]$
		\STATE $\mathcal{G}_{t} =\sqrt{\bar{\alpha}_{t}}\mathcal{G}_{0}+ (1-\bar{\alpha}_{t}) \boldsymbol{\epsilon}$ 
		\STATE $\sigma_{v},\mu_{v} = \phi_{v}(\mathcal{G}_{t}) $
		\STATE Sample $z \sim \mathcal{N}(0, I)$, VAE noise $z_v = \mu_{v}+\sigma_{v}^2z$
		\STATE Regulate $z_v$: \\$\mathcal{L}_{vae}= \mathbb{E}_{q_{\phi}(z_v \mid \mathcal{G}_t)}(-\mathcal{D}_{K L}(q_{\phi}(z_v \mid \mathcal{G}_t)) \| p(z)))$ 
		\STATE Prepare global edges $e_g$ and local edges $e_l$
		\STATE $\boldsymbol{s}_{\theta}\left(\mathcal{G}_{t},z_v, t\right) = \phi_{g}(\mathcal{G}_{t},z_{v}, t, e_g)+\phi_{l}(\mathcal{G}_{t},z_{v}, t, e_l)$
		\STATE Take gradient descent step on\\
		$\nabla_{\theta}\left\|\boldsymbol{s}_{\theta}\left(\mathcal{G}_{t}, z_v, t\right)-\nabla_{\mathcal{G}_{t}} \log q_{\sigma}(\mathcal{G}_{t} \mid \mathcal{G}_{0})\right\|^{2}+\mathcal{L}_{vn,t}$
		\UNTIL Converged 
	\end{algorithmic}  
\end{algorithm}

\subsection{Training}

Having formulated the diffusion and reverse process, the training of the reverse process is performed by optimizing the usual variational lower bound (ELBO) on negative log-likelihood since the exact likelihood is intractable to calculate:
\begin{equation}
\begin{aligned}
&\mathbb{E}\left[-\log p_{\theta}\left(\mathcal{G}\right)\right] \leq \mathbb{E}_{q(\mathcal{G}_0)}[-\log ( \frac{p_{\theta}\left(\mathcal{G}_{0: T}\right)}{q\left(\mathcal{G}_{1: T} \mid \mathcal{G}_{0}\right)})] \\
& =\mathbb{E}_{q(\mathcal{G}_0)}[\underbrace{D_{\mathrm{KL}}\left(q\left(\mathcal{G}_{T} \mid \mathcal{G}_{0}\right) \| p\left(\mathcal{G}_{T}\right)\right)}_{\mathcal{L}_T}\\
&+\sum_{t=2}^{T}\underbrace{D_{\mathrm{KL}}\left(q\left(\mathcal{G}_{t-1} \mid \mathcal{G}_{t}, \mathcal{G}_{0}\right) \| p_{\theta}\left(\mathcal{G}_{t-1} \mid \mathcal{G}_{t}, z_v\right)\right)}_{\mathcal{L}_{t}}\\
&+\sum_{t=2}^{T}\underbrace{\mathbb{E}_{q_{\phi}(z_v \mid \mathcal{G}_t)}(-\mathcal{D}_{K L}(q_{\phi}(z_v \mid \mathcal{G}_t)) \| p(z_v)))}_{\mathcal{L}_{vn,t}}\\
&-\underbrace{\log p_{\theta}\left(\mathcal{G}_{0} \mid \mathcal{G}_{1}\right)}_{\mathcal{L}_0}],
\end{aligned}    
\end{equation}
where $q_\phi(\cdot)$ denotes a learnable variational noising encoder. The detailed derivation is left in the Appendix.

Following \cite{ho2020denoising}, $\mathcal{L}_T$ is a constant and $\mathcal{L}_0$ can be approximated by the product of the PDF of $\mathcal{N}\left(\mathbf{x}_{0} ; \boldsymbol{\mu}_{\theta}\left(\mathbf{x}_{1}, 1\right), \sigma_{1}^{2} I\right)$ and discrete bin width. Hence, we adopt the simplified training objective as follows: 
\begin{equation}
\label{eq: loss_lt}
\mathcal{L}_{t}=\mathbb{E}_{\mathcal{G}_{0}}\left[\gamma\|\boldsymbol{s}_{\theta}\left(\mathcal{G}_{t}, z_v, t\right)
        -\nabla_{\mathcal{G}_{t}} \log q_{\sigma}(\mathcal{G}_{t} \mid \mathcal{G}_{0})\|^{2} \right],
\end{equation}
where $\gamma = \frac{\beta_{t}^{2}}{2 (1-\beta_{t})\left(1-\bar{\alpha}_{t}\right) \sigma_{t}^{2}}$ refers to a weight term.

When $\nabla_{\mathcal{G}_{t}} \log q_{\sigma}(\mathcal{G}_{t} \mid \mathcal{G}_{0})$ denotes a sampling process at stochastic $t$, the sampling on the atomic features $A_t$ still remains invariant. However, the sampling on the atomic coordinates $R_t$ may violate the equivariance. Hence, to maintain the equivariance of $R_t$, we sample the $R_t$ on pairwise distance $\mathbf{d}_{ij}$ instead as
\begin{equation}
    \nabla_{\tilde{\mathbf{r}}_i} \log q_{\sigma}(\tilde{\mathbf{r}}_i \mid \mathbf{r}_i)=\sum_{j \in N(i)} \frac{ \nabla_{\tilde{\mathbf{d}}_{ij}} \log q_{\sigma}(\tilde{\mathbf{d}}_{ij} \mid \mathbf{d}_{ij}) \cdot\left(\mathbf{r}_{i}-\mathbf{r}_{j}\right)}{\mathbf{d}_{ij}},
\end{equation}
where $\tilde{\mathbf{r}}$ denotes the diffused atom coordinate of $\mathcal{G}_t$ and $\tilde{\mathbf{d}}$ denotes the corresponding diffused distance. We approximately calculate $\nabla_{\tilde{\mathbf{d}}} \log q_{\sigma}(\tilde{\mathbf{d}} \mid \mathbf{d})$ as $\frac{-\sqrt{\bar{\alpha}_{t}}(\tilde{\mathbf{d}}-\mathbf{d})}{1-\bar{\alpha}_{t}}$.

Having the aforementioned KL loss of the variational noising, we obtain the final training objectivity:
\begin{equation}
\label{eq: loss_final}
    \mathcal{L} = \sum_{t=2}^{T}\left(\mathcal{L}_{t}+\mathcal{L}_{vn,t}\right). 
\end{equation}

Empirically, if $\gamma$ in Eq. \eqref{eq: loss_lt} is ignored during the training phase, the model performs better instead with the simplified objective. Such simplified objective is equivalent to learning the $\boldsymbol{s}_\theta$ in terms of the gradient of log density of data distribution by sampling the diffused molecule $\mathcal{G}_t$ at a stochastic time step, $t$. 

\begin{algorithm}[t]\small
	\renewcommand{\algorithmicrequire}{\textbf{Input:}}
	\renewcommand{\algorithmicensure}{\textbf{Output:}}
	\caption{Sampling Process}
	\label{alg_s}
	\begin{algorithmic}[1]
	    \REQUIRE The learned global equivariant neural networks $\phi_{g}$, local neural networks $\phi_{l}$
		\ENSURE the molecular coordinates $R$ and atom types $A$
		\FOR{$t=1...T$} 
		\STATE Sample $\mathcal{G}_t \sim \mathcal{N}(0, I)$ 
		\STATE Sample $\xi \sim \mathcal{N}(0, I) \text { if } t>1, \text { else } \xi=\mathbf{0}$
		\STATE Shift $\mathbf{r}_{t}$ to zero COM in $\mathcal{G}_t = \left[r_t, a_t\right]$
		\STATE Prepare global edges $e_g$ and local edges $e_l$
		\STATE Sample $z_{v} \sim \mathcal{N}(0, I)$
		\STATE $\boldsymbol{s}_{\theta}\left(\mathcal{G}_{t}, z_v, t\right) = \phi_{g}(\mathcal{G}_{t}, z_{v}, t, e_g)+\phi_{l}(\mathcal{G}_{t},z_{v}, t, e_l)$
		\STATE $\boldsymbol{\mu}_{\theta}\left(\mathcal{G}_{t}, z_v, t \right)=\frac{1}{\sqrt{1-\beta_{t}}}\left(\mathcal{G}_{t}+\frac{\beta_{t}}{\sqrt{1-\bar{\alpha}_{t}}} \boldsymbol{s}_{\theta}\left(\mathcal{G}_{t}, z_v, t\right)\right)$
		\STATE $\mathcal{G}_{t-1} = \boldsymbol{\mu}_{\theta}\left(\mathcal{G}_{t}, z_v, t\right) +\sigma_{t} \xi$
		\ENDFOR
		\RETURN $\mathcal{G}_0$ to obtain $R$ and $A$
		
	\end{algorithmic}  
\end{algorithm}

Algorithm \ref{alg_t} displays the complete training procedure. Each input molecular with a stochastic time step $t \sim \mathcal{U}(1,T)$ is diffused by the noise $\epsilon$. To ensure the invariance of $\epsilon$, we introduce zero center of mass (COM) from \citeauthor{kohler2020equivariant} (2020) to achieve invariance for $p(\mathcal{G}_T)$. 
By extending the approximation of $p(\mathcal{G}_T)$ from a standard Gaussian to an isotropic Gaussian, the $\epsilon$ is invariant to rotations and translations around the zero COM.

\subsection{Sampling}
To this point, we have the learned reverse Markov kernels $\boldsymbol{s}_\theta$. The mean of the reverse Gaussian transitions $\boldsymbol{\mu}_\theta$ in Eq. \eqref{eq: mu} can be calculated. Figure \ref{fig:model_flow_sampling} illustrates the sampling phase of MDM. Firstly, the chaotic state $\mathcal{G}_T$ is sampled from $\mathcal{N}(0, I)$ and $\boldsymbol{\mu}_\theta$ is obtained by the dual equivariant encoder. The next less chaotic state $\mathcal{G}_{T-1}$ is generated by $\mathcal{N}(\mathcal{G}_T; \boldsymbol{\mu}_\theta, \sigma_T^2I)$. The final molecule $\mathcal{G}_0$ is generated by progressively sample $\mathcal{G}_{t-1}$ for $T$ times. The pseudo code of the sampling process is given in Algorithm \ref{alg_s}.         

\begin{table*}[htb]\small
\centering
\caption{The comparison over 10000 generated molecules of MDM and baseline models on molecular geometry generation task. $\uparrow$ means that higher the values, better the performance of the model.  \label{Tab:2}}
\scalebox{0.8}{
\begin{tabular}{lcccccccc}\\\toprule 
\multirow{2}{*}{Methods} & \multicolumn{4}{c}{QM9} & \multicolumn{4}{c}{GEOM}\\
~ & $\%$ Validity $\uparrow$ & $\%$ Uniqueness $\uparrow$ & $\%$ Novelty $\uparrow$ & $\%$ Stability $\uparrow$ & $\%$ Validity $\uparrow$ & $\%$ Uniqueness $\uparrow$ & $\%$ Novelty $\uparrow$ & $\%$ Stability $\uparrow$\\\midrule
ENF & 41.0 & 40.1 & 39.5 & 24.6 &7.68 & 5.19 & 5.16 & 0\\
G-Schnet & 85.9 & 80.9 & 57.6 & 85.6 & - & - & - & -\\
EDM & 91.7 & 90.5 & 59.9 & 91.1 & 68.6 & 68.6 & 68.6 & 13.7\\
\midrule
MDM-NV & 97.8 & 91.6 & 80.1 & 88.6 & \textbf{99.8} & \textbf{99.5} & \textbf{99.5} & 42.3\\
MDM & \textbf{98.6} & \textbf{94.6} & \textbf{90.0} & \textbf{91.9} & 99.5 & 99.0 & 99.0 & \textbf{62.2}\\
\bottomrule
\end{tabular}}
\end{table*}

\section{Experiments}

In this section, we report the experimental results of the proposed MDM on two benchmark datasets (QM9~\cite{ramakrishnan2014quantum} and GEOM~\cite{axelrod2022geom}), which show that the proposed MDM  significantly outperforms multiple state-of-the-art (SOTA) 3D molecule generation methods. We also conduct additional conditioned generation experiments to evaluate MDM's ability of generating molecules with desired properties. 

\subsection{Molecular Geometry Generation}
\subsubsection{Dataset}
We adopt QM9 \cite{ramakrishnan2014quantum} and GEOM \cite{axelrod2022geom} to evaluate the performance of MDM. QM9 contains over 130K molecules, each containing 18 atoms on average. GEOM contains 290K molecules, each containing 46 atoms on average.
We detail the statistics of two datasets and the corresponding data split setting in the Appendix.

\subsubsection{Baselines and Setup} 
We compare MDM with two one-shot generative models including ENF \cite{satorras2021en} and EDM \cite{hoogeboom2022equivariant}, and one auto-regressive model G-Schnet \cite{gebauer2019symmetry}. 
For ENF and EDM, we utilize their published pre-trained models for evaluation. 
For G-Schnet, we retrain the model on QM9 using its published implementation\footnote{https://github.com/atomistic-machine-learning/G-SchNet}. 
\ul{Note that we did NOT report the results of G-Schnet on GEOM dataset because its published implementation does not provide the data process scripts and the corresponding configuration files for GEOM dataset.}
In addition, we introduce `MDM-NV' (\textbf{N}o \textbf{V}ariational), which excludes the controlling variable $z_v$ and $\mathcal{L}_{vn}$  from MDM, to study their impact. 

For all the scenarios in this section, we use 10000 generated samples for evaluation. The molecules generated from QM9 include all kinds of chemical bonds. Since the molecules in the GEOM dataset are quite large and the structure is very complex. It is hard to build the chemical bond via the atomic pairwise distances. Hence, we only consider building single bonds to generate the molecules. \footnote{We also report the results of building all kinds of chemical bonds in the Appendix.} 

\subsubsection{Metrics}

We measure the generation performance via four metrics: 
\begin{itemize}\small
    \item \textbf{Validity}: the percentage of the generated molecules that follow the chemical valency rules specified by RDkit; 
    \item \textbf{Uniqueness}: the percentage of unique \& valid molecules in all the generated samples;
    \item \textbf{Novelty}: the percentage of generated unique molecules that are not in the training set; 
    \item \textbf{Stability}: the percentages of the generated molecules that do not include ions.\footnote{The existance of ions indicates that there are two molecular fragments in the generated molecule without bond connection.}
\end{itemize}

\begin{figure*}[ht!]
    \centering
    \includegraphics[width=1\textwidth, height=0.15\textwidth]{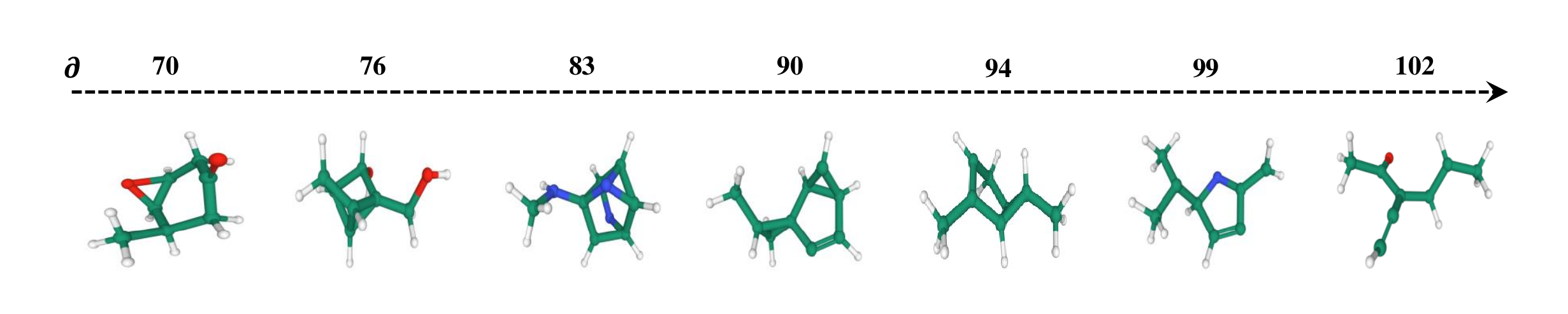}
    \caption{Generated molecules with different Polarizability $\alpha$ by conditional MDM with the same initial noise. The $\alpha$ values of each generated molecule are presented on top of the dash line.}
    \label{fig:quant_alpha_1}
\end{figure*}

\subsection{Results and Analysis}

In Table~\ref{Tab:2}, we report the performances of all the models in terms of four metrics on both QM9 and GEOM datasets. 
From Table~\ref{Tab:2}, we can see that the proposed MDM and its variant outperform all the baseline models. 
For instance, on the QM9 dataset, MDM defeats SOTA (i.e., EDM) by $6.9\%$ in Validity, $4.1\%$ in Uniqueness and $31.1\%$ in Novelty. 
On GEOM dataset, the performance gaps even increase to $30.9\%$ in Validity, $30.4\%$ in Uniqueness, $30.4\%$ in Novelty, and $48.5\%$ in Stability.
By outperforming various SOTA generation approaches, the proposed MDM demonstrates its advantage of generating high-quality molecules, especially when the generated molecules contain a larger number of atoms on average (GEOM v.s. QM9).

The reason behind such significant improvements is that MDM involves two independent equivalent graph networks to discriminately considers two types of inter-atomic forces, i.e., chemical bonds and van der Waals forces. The huge strength difference between these two types of forces leads to significant distinct local geometries between neighbor atoms with different atomic spacing, especially for larger molecules with more atoms and more complex local geometry structures (GEOM v.s. QM9). 
Such a justification is further supported by the fact that MDM-NV also outperforms EDM, given that both models utilize a diffusion-based framework as the backbone. 
In contrast to EDM, MDM and its variant MDM-NV successfully capture structural patterns of different atomic spacing and generate molecules that highly follow chemical valency rules (high validity) and few ions (low stability).

Besides, we also witness that MDM achieves salient improvements in uniqueness and novelty compared with its variant MDM-NV. Such improvements indicate that the controlling variable $z$ provides thorough explorations in the intermediate steps of the Langevin dynamics and further leads to more diverse generation results.

\subsection{Conditional Molecular Generation}
\subsubsection{Baselines and Setup}
In this section, we present the conditional molecular generation in which
we train our model conditioned with six properties Polarizability $\alpha$, HOMO $\epsilon_{\text{HOMO}}$, LUMO $\epsilon_{\text{LUMO}}$, HOMO-LUMO gap $\epsilon_{\text{gap}}$, Dipole moment $\mu$ and $C_v$ on QM9 dataset. Here, we implement the conditioned generation by concatenating the property values $c$ with the atom features to obtain $p_\theta(\mathcal{G}_{t-1} \mid \mathcal{G}_t, c)$. 

Following previous work \cite{hoogeboom2022equivariant}, we utilize the property classifier from \cite{satorras2021en}. The training set of QM9 is divided into two halves each containing 50K samples. One half  ($D_t$) is used for classifier training, and the other one ($D_e$) is utilized for generative models training. 
Then the classifier $\phi_c$ evaluates the conditional generated samples by Mean Absolute Error (MAE) of the predicted and true property values.

\subsubsection{Baselines}
Here, we provide several baseline references for comparisons:
\begin{itemize}\small
    \item \textbf{Naive (Upper-Bond)}: the classifier $\phi_c$ evaluates on $D_e$ in which the labels are shuffled to predict the molecular properties. This is the upper-bound of possible MAEs (the worst case). 
    \item \textbf{\#Atoms}: the classifier $\phi_c$ only depends on the number of atoms to predict the molecular properties on $D_e$.
    \item \textbf{QM9 (Lower-Bond)}: the classifier directly evaluates on original $D_t$ to predict the molecular properties. 
\end{itemize}

Given the MAE score for samples generated by a generative model, \ul{the smaller gap between its MAE score and ``QM9 (Lower-Bond)'', the better corresponding model fits the the data distribution on $D_e$}.
Suppose the MAE score of a model outperforms ``\#Atoms''. In that case, it suggests that the model incorporates the targeted property information in the generation process instead of simply generating samples with a specific number of atoms.   

\begin{table}[htb]\small
\centering
\caption{The results of conditional molecular generation on QM9 dataset. $\downarrow$ means the lower the values, the better the model incorporates the targeted properties. \label{Tab:3}}
\scalebox{0.8}{
{\begin{tabular}{lcccccc} \\\toprule  
Methods & $\alpha$ $\downarrow$ & $\epsilon_{\text{{gap}}}$ $\downarrow$ & $\epsilon_{\text{{HOMO}}}$ $\downarrow$ & $\epsilon_{\text{{LUMO}}}$ $\downarrow$ & $\mu$ $\downarrow$& $C_v$ $\downarrow$ \\\midrule
Naive (U-bound) & 9.013 & 1.472 & 0.645 & 1.457 & 1.616 & 6.857\\
\#Atoms & 3.862 & 0.866 & 0.426 & 0.813 & 1.053 & 1.971\\
\midrule
EDM & 2.760 & 0.655 & 0.356 & 0.584 & 1.111 & 1.101\\
MDM & 1.591 & 0.044 & 0.019 & 0.040 & 1.177 & 1.647\\
\midrule
QM9(L-bound) & 0.100 & 0.064 & 0.039 & 0.036 & 0.043 & 0.040\\\bottomrule
\end{tabular}}}
\end{table}

\subsubsection{Results}

Table \ref{Tab:3} presents the targeted generation results.  In particular, MDM surpasses "Naive", "\#Atoms", and EDM in almost all the properties except for $\mu$ and $C_v$. The results indicate that MDM performs better than EDM in incorporating the targeted property information into the generated samples themselves beyond the number of features. Moreover, we notice that the MAE of MDM on gaps and HOMO is even slightly lower than "QM9 (Lower-Bound)". 
This phenomenon may be caused by the slight distribution difference between $D_e$ and $D_t$. At the same time, it indicates that MDM can well fit the distribution of $D_e$ and generate high-quality molecules with targeted properties.  
Empirically, the conditional generation will not hurt the quality of the generated molecules in terms of validity, uniqueness, novelty, and stability.


Apart from the quantitative analysis, we also provide case studies to analyze the effect of applying different values of the properties to conditional generation. Here, we adopt the property Polarizability $\alpha$ for demonstration. Generally, a molecule with a larger Polarizability is expected to be less isometrically shaped. We fix the number of atoms as 19 since it is the most frequent molecular size in QM9. Figure \ref{fig:quant_alpha_1} outlines the molecules generated by conditional MDM when interpolating different Polarizability values, which is in line with our expectation.

\section{Conclusion}
In this study, we propose a novel diffusion model MDM to generate 3D molecules from scratch. MDM augments interatomic linkages that are not in the 3D point cloud representation of molecules and proposes separated equivariant encoders to capture the interatomic forces of different strengths. In addition, we introduce a controlling variable in both diffusion and reverse processes to improve generation diversity. Comprehensive experiments demonstrate that MDM exceeds previous SOTA models by a non-trivial margin and can generate molecules with desired properties. 

\bibliography{aaai22}

\newpage
\onecolumn

\section{Notations used in this paper}
We provide notations used in this paper for easier reading.
\begin{table}[htbp]\small%
\centering
\begin{tabular}{l|l}
Notations & Descriptions \\
\hline
$t$ & Time step \\
\hline
$A$ & The atom representation\\
\hline
$R$ & The position of the atoms \\
\hline
$\mathcal{G}_0$ & The ground truth molecule geometry\\
\hline
$\mathcal{G}_{1...T}$ & The latent information via diffusion\\
\hline
$\beta_t$ & A fixed variance schedule\\
\hline
$\alpha_t$ & $\alpha_t = 1- \beta_t$\\
\hline
$\bar{\alpha}_t$ & $\bar{\alpha}_{t}=\prod_{s=1}^{t} 1-\beta_{s}$\\
\hline
$\mathbf{d}_{ij}$ & The Euclidean distance between atom $i$ and atom $j$\\
\hline
$e_{ij}$ & Edge between atom $i$ and atom $j$\\
\hline
$\sigma_t^2$ & User defined variance \\
\hline
$\sigma_v$ & The standard deviation outputted by VAE encoder\\
\hline
$\mu_v$ & The mean outputted by VAE encoder\\
\hline
$\boldsymbol{\epsilon}_{\theta}$ & Parameterized noise \\
\hline
$\boldsymbol{\epsilon}$ & $\boldsymbol{\epsilon} = \left[\boldsymbol{\epsilon}^a, \boldsymbol{\epsilon}^r \right]$ where $\boldsymbol{\epsilon}^a \sim \mathcal{N}(0, I)$,$\boldsymbol{\epsilon}^r \sim \mathcal{N}(0, I)$ \\
\hline
$\boldsymbol{s}_{\theta}$ & Parameterized stein score\\
\hline
$\boldsymbol{\mu}_{\theta}$ & Parameterized mean\\
\hline
$z_{v}$ & VAE noise\\
\hline
$z$ & Sample from $\mathcal{N}(0, I)$ \\
\hline
$\xi$ & Sample from $\mathcal{N}(0, I)$ \\
\hline
$\bar{\epsilon}$ & Sample from $\mathcal{N}(0, I)$ \\
\hline
$\phi$ & Neural networks \\
\hline
$q()$ & Distribution of diffusion process \\
\hline
$p()$ & Distribution of reverse process \\
\hline
$c$ & The conditional property \\
\hline
$[,]$ & The concatenation operation\\
\hline
$\odot$ & Hadamard product
\end{tabular}
\caption{Notations used in this paper.}
\label{table1}
\end{table}
\section{Proof of the diffusion model}

We provide proofs for the derivation of several properties in the diffusion model. For the detailed explanation and discussion, we refer readers to \cite{ho2020denoising}.  

\subsection{Marginal distribution of the diffusion process}
In the diffusion process, we have the marginal distribution of the data at any arbitrary time step $t$ in a closed form:
\begin{equation}
    \label{eq: q of any arbitrary time step}
q\left(\mathcal{G}_{t} \mid \mathcal{G}_{0}\right)=\mathcal{N}\left(\mathcal{G}_{t} ; \sqrt{\bar{\alpha}_{t}} \mathcal{G}_{0},\left(1-\bar{\alpha}_{t}\right) I\right).
\end{equation}

Recall the posterior $q\left(\mathcal{G}_{t} \mid \mathcal{G}_{0}\right)$ in Eq.2 (main document), we can obtain $\mathcal{G}_{t}$ using the reparameterization trick. A property of the Gaussian distribution is that if we add $\mathcal{N}(0,  \sigma^2_1I)$ and $\mathcal{N}(0, \sigma^2_2I)$, the new distribution is $\mathcal{N}(0,  (\sigma^2_1+\sigma^2_2)I)$
\begin{equation}
\label{eq: derivation of maginal distribution}
    \begin{split}
        \mathcal{G}_{t} &= \sqrt{\alpha_{t}}\mathcal{G}_{t-1} + \sqrt{1-\alpha_{t}}\epsilon_{t-1}\\
        &=\sqrt{\alpha_{t}\alpha_{t-1}}\mathcal{G}_{t-2} + \sqrt{\alpha_{t}(1-\alpha_{t-1})}\epsilon_{t-2} + \sqrt{1-\alpha_{t}}\epsilon_{t-1}\\
        &=\sqrt{\alpha_{t}\alpha_{t-1}}\mathcal{G}_{t-2} + \sqrt{1-\alpha_t\alpha_{t-1}}\bar{\epsilon}_{t-2}\\
        &=\dots\\
        &=\sqrt{\bar{\alpha}_t}\mathcal{G}_{0} + \sqrt{1-\bar{\alpha_t}}\bar{\epsilon},
    \end{split}
\end{equation}
where $\alpha_t=1-\beta_t$, $\epsilon$ and $\hat{\epsilon}$ are sampled from independent standard Gaussian distributions.

\subsection{The parameterized mean $\mathbf{\mu}_{\theta}$}
A learned Gaussian transitions $p_{\theta}\left(\mathcal{G}_{t-1} \mid  \mathcal{G}_{t}\right)$ is devised to approximate the $q\left(\mathcal{G}_{t-1} \mid  \mathcal{G}_{t}\right)$ of every time step: $p_{\theta}\left(\mathcal{G}_{t-1} \mid  \mathcal{G}_{t}\right)=\mathcal{N}\left(\mathcal{G}_{t-1} ; \boldsymbol{\mu}_{\theta}\left(\mathcal{G}_{t}, t\right), \sigma_{t}^{2} I\right)$. $\boldsymbol{\mu}_\theta$ is parameterized as follows:
\begin{equation}
    \boldsymbol{\mu}_{\theta}\left(\mathcal{G}_{t}, t\right)=\frac{1}{\sqrt{\alpha_{t}}}\left(\mathcal{G}_{t}-\frac{\beta_{t}}{\sqrt{1-\bar{\alpha}_{t}}} \boldsymbol{\epsilon}_{\theta}\left(\mathcal{G}_{t}, t\right)\right).
\end{equation}

 The distribution $q\left(\mathcal{G}_{t-1} \mid  \mathcal{G}_{t}\right)$ can be expanded by Bayes' rule:
\begin{equation}
\begin{split}
    q\left(\mathcal{G}_{t-1} \mid  \mathcal{G}_{t}\right) &= q\left(\mathcal{G}_{t-1} \mid  \mathcal{G}_{t}, \mathcal{G}_{0}, \right)\\
    &=q\left(\mathcal{G}_{t} \mid \mathcal{G}_{t-1}, \mathcal{G}_{0}\right) \frac{q\left(\mathcal{G}_{t-1} \mid \mathcal{G}_{0}\right)}{q\left(\mathcal{G}_{t} \mid \mathcal{G}_{0}\right)} \\
    &=q\left(\mathcal{G}_{t} \mid \mathcal{G}_{t-1}\right) \frac{q\left(\mathcal{G}_{t-1} \mathcal{G}_{0}\right)}{q\left(\mathcal{G}_{t} \mid \mathcal{G}_{0}\right)} \\
    &\propto \exp \left(-\frac{1}{2}\left(\frac{\left(\mathcal{G}_{t}-\sqrt{\alpha_{t}} \mathcal{G}_{t-1}\right)^{2}}{\beta_{t}}+\frac{\left(\mathcal{G}_{t-1}-\sqrt{\alpha_{t-1}} \mathcal{G}_{0}\right)^{2}}{1-\bar{\alpha}_{t-1}}-\frac{\left(\mathcal{G}_{t}-\sqrt{\alpha_{t}} \mathcal{G}_{0}\right)^{2}}{1-\bar{\alpha}_{t}}\right)\right) \\
    &=\exp \left(-\frac{1}{2}\left(\left(\frac{\alpha_{t}}{\beta_{t}}+\frac{1}{1-\bar{\alpha}_{t-1}}\right) \mathcal{G}_{t-1}^{2}-\left(\frac{2 \sqrt{\alpha_{t}}}{\beta_{t}} \mathcal{G}_{t}+\frac{2 \sqrt{\alpha_{t-1}}}{1-\bar{\alpha}_{t-1}} \mathcal{G}_{0}\right) \mathcal{G}_{t-1}+C\left(\mathcal{G}_{t}, \mathcal{G}_{0}\right)\right)\right)\\
    &\propto \exp (-\mathcal{G}_{t-1}^{2} + (\frac{\sqrt{\alpha_{t}}\left(1-\bar{\alpha}_{t-1}\right)}{1-\bar{\alpha}_{t}}\mathcal{G}_{t}+\frac{\sqrt{\bar{\alpha}_{t-1}} \beta_{t}}{1-\bar{\alpha}_{t}}\mathcal{G}_{0})\mathcal{G}_{t-1}),
\end{split}
\end{equation}
where $C\left(\mathcal{G}_{t}, \mathcal{G}_{0}\right)$ is a constant. We can find that $q\left(\mathcal{G}_{t-1} \mid  \mathcal{G}_{t}\right)$ is also a Gaussian distribution. We assume that:
\begin{equation}
    q\left(\mathcal{G}_{t-1} \mid \mathcal{G}_{t}, \mathcal{G}_{0}\right)=\mathcal{N}\left(\mathcal{G}_{t-1} ; \tilde{\boldsymbol{\mu}}\left(\mathcal{G}_{t}, \mathcal{G}_{0}\right), \tilde{\beta}_{t} I\right),
\end{equation}
where $\tilde{\beta}_{t}=1 /\left(\frac{\alpha_{t}}{\beta_{t}}+\frac{1}{1-\bar{\alpha}_{t-1}}\right)=\frac{1-\bar{\alpha}_{t-1}}{1-\bar{\alpha}_{t}} \cdot \beta_{t}$ and $\tilde{\boldsymbol{\mu}}_{t}\left(\mathcal{G}_{t}, \mathcal{G}_{0}\right)=\left(\frac{\sqrt{\alpha_{t}}}{\beta_{t}} \mathcal{G}_{t}+\frac{\sqrt{\bar{\alpha}_{t-1}}}{1-\bar{\alpha}_{t-1}} \mathcal{G}_{0}\right) /\left(\frac{\alpha_{t}}{\beta_{t}}+\frac{1}{1-\bar{\alpha}_{t-1}}\right)=\frac{\sqrt{\alpha_{t}}\left(1-\bar{\alpha}_{t-1}\right)}{1-\bar{\alpha}_{t}} \mathcal{G}_{t}+\frac{\sqrt{\bar{\alpha}_{t-1}} \beta_{t}}{1-\bar{\alpha}_{t}} \mathcal{G}_{0}$.

From Eq. \ref{eq: derivation of maginal distribution}, we have $\mathcal{G}_{t}==\sqrt{\bar{\alpha}_t}\mathcal{G}_{0} + \sqrt{1-\bar{\alpha_t}}\bar{\epsilon}$. We take this into $\tilde{\boldsymbol{\mu}}$:
\begin{equation}
    \begin{aligned}
\tilde{\boldsymbol{\mu}}_{t} &=\frac{\sqrt{\alpha_{t}}\left(1-\bar{\alpha}_{t-1}\right)}{1-\bar{\alpha}_{t}} \mathbf{x}_{t}+\frac{\sqrt{\bar{\alpha}_{t-1}} \beta_{t}}{1-\bar{\alpha}_{t}} \frac{1}{\sqrt{\bar{\alpha}_{t}}}\left(\mathbf{x}_{t}-\sqrt{1-\bar{\alpha}_{t}} \mathbf{\epsilon}_{t}\right) \\
&=\frac{1}{\sqrt{\alpha_{t}}}\left(\mathbf{x}_{t}-\frac{\beta_{t}}{\sqrt{1-\bar{\alpha}_{t}}} \mathbf{\epsilon}_{t}\right).
\end{aligned}
\end{equation}

$\boldsymbol{\mu}_\theta$ is designed to model $\tilde{\boldsymbol{\mu}}$. Therefore, $\boldsymbol{\mu}_\theta$ has the same formulation as $\tilde{\boldsymbol{\mu}}$ but parameterizes $\epsilon$:
\begin{equation}
    \boldsymbol{\mu}_{\theta}\left(\mathcal{G}_{t}, t\right)=\frac{1}{\sqrt{\alpha_{t}}}\left(\mathcal{G}_{t}-\frac{\beta_{t}}{\sqrt{1-\bar{\alpha}_{t}}} \boldsymbol{\epsilon}_{\theta}\left(\mathcal{G}_{t}, t\right)\right).
\end{equation}

\subsection{Decompose atomic coordinates to pairwise distances}
In order to achieve the equivariance of the atomic coordinates in 3D space, we attempt to decompose them to pairwise distances.

\begin{equation}
    g(R) = \mathbf{d},
\end{equation}
\begin{equation}
    \log p_{\theta}(R) \triangleq f \circ g(R)=f(\mathbf{d}),
\end{equation}
where $g: \mathbb{R}^{n \times 3} \rightarrow \mathbb{R}^{|E| \times 1}$ denotes a function that maps the $n$ atomic coordinates to $|E|$ interatomic distances and $\mathbb{R}^{|E|} \rightarrow \mathbb{R}$ is a neural network that estimates the log density of a molecule based on the interatomic distances $\mathbf{d}$.

\begin{equation}
\begin{aligned}
\forall i, \mathbf{s}_{\theta}(R_{i}) &=\frac{\partial f(\mathbf{d})}{\partial \mathbf{r}_{i}}=\sum_{(i, j), e_{i j} \in E} \frac{\partial f(\mathbf{d})}{\partial \mathbf{d}_{i j}} \cdot \frac{\partial \mathbf{d}_{i j}}{\partial \mathbf{r}_{i}} \\
&=\sum_{j \in N(i)} \frac{1}{\mathbf{d}_{i j}} \cdot \frac{\partial f(\mathbf{d})}{\partial \mathbf{d}_{i j}} \cdot\left(\mathbf{r}_{i}-\mathbf{r}_{j}\right) \\
&=\sum_{j \in N(i)} \frac{1}{\mathbf{d}_{i j}} \cdot \mathbf{s}_{\theta}(\mathbf{d}_{i j}) \cdot\left(\mathbf{r}_{i}-\mathbf{r}_{j}\right).
\end{aligned}   
\end{equation}

We refer readers to \cite{shi2021learning} for details.

\subsection{The ELBO objective}

It is hard to directly calculate log likelihood of the data. Instead, we can derive its ELBO objective for optimizing.  
\begin{equation}
\begin{split}
    \mathbb{E}\left[-\log p_{\theta}\left(\mathcal{G}\right)\right]
    &=-\mathbb{E}_{q\left(\mathcal{G}_{0}\right)} \log \left(\int p_{\theta}\left(\mathcal{G}_{0: T}, z_v\right) d \mathcal{G}_{1: T}\right) \\
    &=-\mathbb{E}_{q\left(\mathcal{G}_{0}\right)} \log \left(\int q\left(\mathcal{G}_{1: T} \mid \mathcal{G}_{0}\right) \frac{p_{\theta}\left(\mathcal{G}_{0: T}, z_v\right)}{q\left(\mathcal{G}_{1: T} \mid \mathcal{G}_{0}\right)} d \mathcal{G}_{1: T}\right) \\
    &=-\mathbb{E}_{q\left(\mathcal{G}_{0}\right)} \log \left(\mathbb{E}_{q\left(\mathcal{G}_{1: T} \mid \mathcal{G}_{0}\right)} \frac{p_{\theta}\left(\mathcal{G}_{0: T}, z_v\right)}{q\left(\mathcal{G}_{1: T} \mid \mathcal{G}_{0}\right)}\right) \\
    &\leq-\mathbb{E}_{q\left(\mathcal{G}_{0: T}\right)} \log \frac{p_{\theta}\left(\mathcal{G}_{0: T}, z_v\right)}{q\left(\mathcal{G}_{1: T} \mid \mathcal{G}_{0}\right)} \\
    &=\mathbb{E}_{q\left(\mathcal{G}_{0: T}\right)}\left[\log \frac{q\left(\mathcal{G}_{1: T} \mid \mathcal{G}_{0}\right)}{p_{\theta}\left(\mathcal{G}_{0: T}, z_v\right)}\right].
\end{split}
\end{equation}

Then we further derive the ELBO objective:
\begin{equation}
    \begin{split}
        \mathbb{E}_{q\left(\mathcal{G}_{0: T}\right)}\left[\log \frac{q\left(\mathcal{G}_{1: T} \mid \mathcal{G}_{0}\right)}{p_{\theta}\left(\mathcal{G}_{0: T}, z_v\right)}\right]
&=\mathbb{E}_{q}\left[\log \frac{\prod_{t=1}^{T} q\left(\mathcal{G}_{t} \mid \mathcal{G}_{t-1}\right)}{p_{\theta}\left(\mathcal{G}_{T}, z_v\right) \prod_{t=1}^{T} p_{\theta}\left(\mathcal{G}_{t-1} \mid \mathcal{G}_{t}, z_v\right)}\right]\\
&=\mathbb{E}_{q}\left[-\log p_{\theta}\left(\mathcal{G}_{T}, z_v\right)+\sum_{t=1}^{T} \log \frac{q\left(\mathcal{G}_{t} \mid \mathcal{G}_{t-1}\right)}{p_{\theta}\left(\mathcal{G}_{t-1} \mid \mathcal{G}_{t}, z_v\right)}\right]\\
&=\mathbb{E}_{q}\left[-\log p_{\theta}\left(\mathcal{G}_{T}, z_v\right)+\sum_{t=2}^{T} \log \frac{q\left(\mathcal{G}_{t} \mid \mathcal{G}_{t-1}\right)}{p_{\theta}\left(\mathcal{G}_{t-1} \mid \mathcal{G}_{t}, z_v\right)}+\log \frac{q\left(\mathcal{G}_{1} \mid \mathcal{G}_{0}\right)}{p_{\theta}\left(\mathcal{G}_{0} \mid \mathcal{G}_{1}, z_v\right)}\right]\\
&=\mathbb{E}_{q}\left[-\log p_{\theta}\left(\mathcal{G}_{T}, z_v\right)+\sum_{t=2}^{T} \log \left(\frac{q\left(\mathcal{G}_{t-1} \mid \mathcal{G}_{t}, \mathcal{G}_{0}\right)}{p_{\theta}\left(\mathcal{G}_{t-1} \mid \mathcal{G}_{t}, z_v\right)} \cdot \frac{q\left(\mathcal{G}_{t} \mid \mathcal{G}_{0}\right)}{q\left(\mathcal{G}_{t-1} \mid \mathcal{G}_{0}\right)}\right)+\log \frac{q\left(\mathcal{G}_{1} \mid \mathcal{G}_{0}\right)}{p_{\theta}\left(\mathcal{G}_{0} \mid \mathcal{G}_{1}, z_v\right)}\right]\\
&=\mathbb{E}_{q}\left[-\log p_{\theta}\left(\mathcal{G}_{T}, z_v\right)+\sum_{t=2}^{T} \log \frac{q\left(\mathcal{G}_{t-1} \mid \mathcal{G}_{t}, \mathcal{G}_{0}\right)}{p_{\theta}\left(\mathcal{G}_{t-1} \mid \mathcal{G}_{t}, z_v\right)}+\sum_{t=2}^{T} \log \frac{q\left(\mathcal{G}_{t} \mid \mathcal{G}_{0}\right)}{q\left(\mathcal{G}_{t-1} \mid \mathcal{G}_{0}\right)}+\log \frac{q\left(\mathcal{G}_{1} \mid \mathcal{G}_{0}\right)}{p_{\theta}\left(\mathcal{G}_{0} \mid \mathcal{G}_{1}, z_v\right)}\right]\\
&=\mathbb{E}_{q}\left[-\log p_{\theta}\left(\mathcal{G}_{T}, z_v\right)+\sum_{t=2}^{T} \log \frac{q\left(\mathcal{G}_{t-1} \mid \mathcal{G}_{t}, \mathcal{G}_{0}\right)}{p_{\theta}\left(\mathcal{G}_{t-1} \mid \mathcal{G}_{t}, z_v\right)}+\log \frac{q\left(\mathcal{G}_{T} \mid \mathcal{G}_{0}\right)}{q\left(\mathcal{G}_{1} \mid \mathcal{G}_{0}\right)}+\log \frac{q\left(\mathcal{G}_{1} \mid \mathcal{G}_{0}\right)}{p_{\theta}\left(\mathcal{G}_{0} \mid \mathcal{G}_{1}, z_v\right)}\right]\\
&=\mathbb{E}_{q}\left[\log \frac{q\left(\mathcal{G}_{T} \mid \mathcal{G}_{0}\right)}{p_{\theta}\left(\mathcal{G}_{T}, z_v\right)}+\sum_{t=2}^{T} \log \frac{q\left(\mathcal{G}_{t-1} \mid \mathcal{G}_{t}, \mathcal{G}_{0}\right)}{p_{\theta}\left(\mathcal{G}_{t-1} \mid \mathcal{G}_{t}, z_v\right)}-\log p_{\theta}\left(\mathcal{G}_{0} \mid \mathcal{G}_{1}, z_v\right)\right]\\
&=\mathbb{E}_{q}\Bigg{[}\underbrace{D_{\mathrm{KL}}\left(q\left(\mathcal{G}_{T} \mid \mathcal{G}_{0}\right) \| p_{\theta}\left(\mathcal{G}_{T}, z_v\right)\right)}_{L_{T}}+\\
&\sum_{t=2}^{T} \underbrace{D_{\mathrm{KL}}\left(q\left(\mathcal{G}_{t-1} \mid \mathcal{G}_{t}, \mathcal{G}_{0}\right) \| p_{\theta}\left(\mathcal{G}_{t-1} \mid \mathcal{G}_{t}, z_v\right)\right)}_{L_{t}}\underbrace{-\log p_{\theta}\left(\mathcal{G}_{0} \mid \mathcal{G}_{1}, z_v\right)}_{L_{0}}\bigg{]}.
    \end{split}
\end{equation}

\section{Experiments details}
\subsection{Dataset and implementation}
\subsubsection{QM9}
QM9 dataset contains over 130K small molecules with quantum chemical properties which each consist of up to 9 heavy atoms or 29 atoms including hydrogens. On average, each molecule contains 18 atoms. For a fair comparison, we follow the previous work \cite{anderson2019cormorant} to split the data into training, validation and test set, which each partition contains 100K, 18K and 13K molecules respectively.

MDM is trained by Adam \cite{DBLP:journals/corr/KingmaB14} optimizer for 200K iterations (about 512 epochs) with a batch size of 256 and a learning rate of 0.001.

\subsubsection{Geom}
Following previous work \cite{hoogeboom2022equivariant}, we evaluate MDM on a larger scale dataset GEOM \cite{axelrod2022geom}. Compared to QM9, the size of molecules in GEOM is much larger, in which is up to 181 atoms and 46 atoms on average (including hydrogens). We obtain the lowest energy conformation for each molecule, and finally we have 290K samples for training.

MDM is trained by Adam \cite{DBLP:journals/corr/KingmaB14} optimizer for 200K iterations (about 170 epochs) with a batch size of 256 and a learning rate of 0.001.

\subsection{Bond prediction}
Since we only have the atom types and atom coordinates as the output, we need to predict the bonds according to the atomic pairwise distances. In this paper, we follow \cite{hoogeboom2022equivariant} to construct the bonds. If the atomic pairwise distances are within the specified range according to the chemical bonds lookup tables \footnote{http://chemistry-reference.com/tables/Bond\%20Lengths\%20and\%20Enthalpies.pdf}, then we add the corresponding bonds.

For example, we add a single bond for the atom $i$ and atom $j$ if 
$D_{d, ij} \leq \mathbf{d}_{ij} \leq D_{s,ij}$, 
where $D_{d, ij}$ denotes the referenced double bonds length and $D_{s, ij}$ denotes the referenced double bonds length. 

\subsection{Selected properties}
In this paper, we select six properties to evaluate the conditional generation performance of MDM.
\begin{itemize}
    \item \textbf{Polarizability $\alpha$}: the response of electron distribution of a molecule when subjected to an externally-applied static electric field.
    \item \textbf{HOMO $\epsilon_{\text{HOMO}}$}: the energy of the highest occupied molecular orbit.
    \item \textbf{LUMO $\epsilon_{\text{LUMO}}$}: the energy of the lowest occupied molecular orbit.
    \item \textbf{HOMO-LUMO gap $\epsilon_{\text{gap}}$}: the energy difference between the HOMO and LUMO.
    \item \textbf{Dipole moment $\mu$}: measurement of the molecular electric dipole moment.
    \item \textbf{$C_v$}: the heat capacity at 298.15K. 
\end{itemize}

\subsection{Additional experiment results on Geom}
The molecules in GEOM dataset are quite large, and the structure is very complex, EDM \cite{hoogeboom2022equivariant} cannot build the chemical bonds via the atomic pairwise distances. Thus, it only considers building single bonds to generate the molecules. However, we should consider all kinds of chemical bonds such as double and triple bonds to build the molecule. Here, we report the results of molecules generated by different models on Geom which include all kinds of chemical bonds.  

\begin{table}[htb]
\centering
\caption{The comparison over 10000 generated molecules which include all kinds of chemical bonds of different models on GEOM dataset. $\uparrow$ means that higher the values, better the performance of the model. \label{Tab:single bond result on QM9}}
\scalebox{1}{
{\begin{tabular}{lcccc} \\\toprule  
Methods & $\%$ Validity $\uparrow$ & $\%$ Uniqueness $\uparrow$ & $\%$ Novelty $\uparrow$ & $\%$ Stability $\uparrow$ \\\midrule
ENF & 7.9 & 5.4 & 5.4 & 0\\
EDM & 0.19 & 0.19 & 0.19 & 0.03\\
MDM (no VAE) & 94.4 & 94.2 & 93.8 & 40.0\\
MDM & \textbf{95.6} & \textbf{94.9} & \textbf{94.6} & \textbf{61.9}\\\midrule
\bottomrule
\end{tabular}}}{}
\end{table}

As shown in Table \ref{Tab:single bond result on QM9}, we observe that the performance of EDM drops significantly on all metrics, compared to considering single chemical bonds. On the contrary, our model still performs well in this setting and surpasses all the baseline models by a significant margin.

\section{Samples generated by MDM}
\begin{figure}[htbp]
    \centering
    \includegraphics[width=\linewidth]{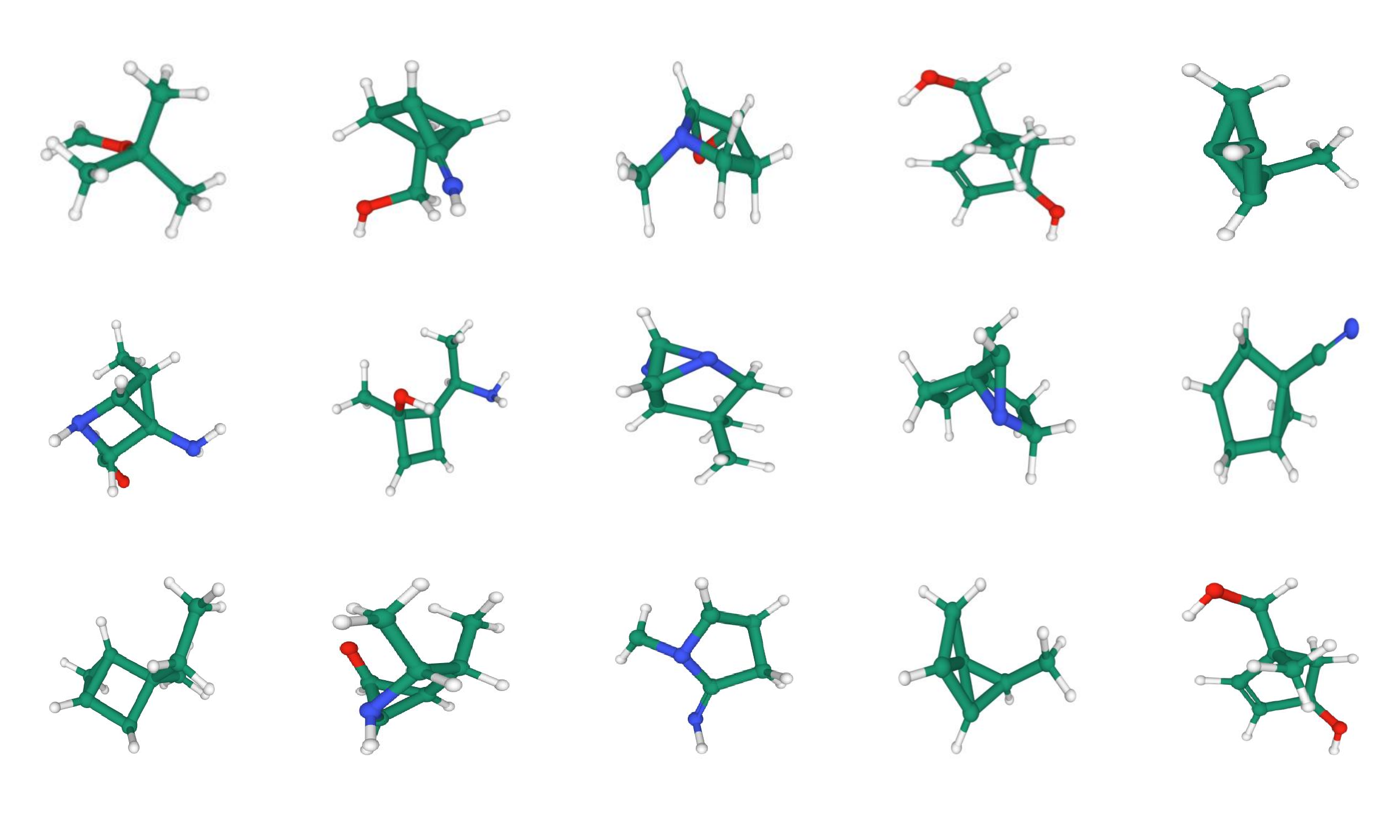}
    \caption{Generated Samples on QM9}
    \label{fig:QM9 samples}
    \includegraphics[width=\linewidth]{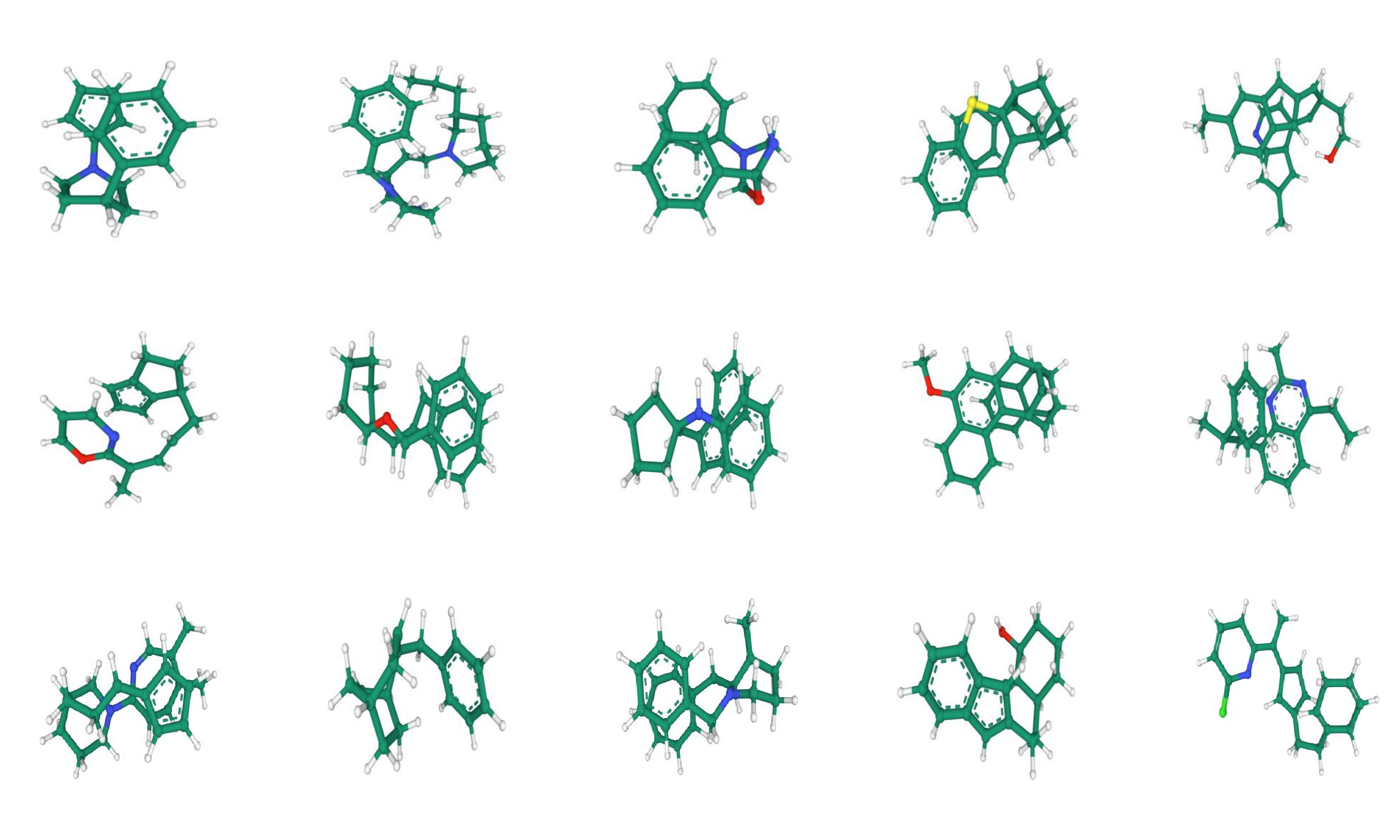}
    \caption{Generated Samples on Geom}
    \label{fig:Geom samples}
\end{figure}

We provide more visualizations of generated samples which are trained on QM9 and Geom in Figure \ref{fig:QM9 samples} and Figure \ref{fig:Geom samples}.

\end{document}